# Controlled Language and Baby Turing Test for General Conversational Intelligence

Anton Kolonin[1,2,3]

[1]Aigents, Novosibirsk, Russian Federation
[2]SingularityNET Foundation, Amsterdam, Netherlands
[3]Novosibirsk State University, Novosibirsk, Russia
akolonin@gmail.com

**Abstract.** General conversational intelligence appears to be an important part of artificial general intelligence. Respectively, it requires accessible measures of the intelligence quality and controllable ways of its achievement, ideally – having the linguistic and semantic models represented in a reasonable way. Our work is suggesting to use "Baby Turing Test" approach to extend the classic "Turing Test" for conversational intelligence and controlled language based on semantic graph representation extensible for arbitrary subject domain. We describe how the two can be used together to build a general-purpose conversational system such as intelligent assistant for online media and social network data processing.

## 1    Introduction

The modern state of conversational intelligence [1,2] can be seen as a field where either one of the two kinds of natural language processing system is present: A) dialog flow-based domain-specific application serving specific intents under scope of manually pre-defined domains [3]; B) neural network-based systems for free-style chat trained on past conversations [4]. The most advanced systems combine both approaches as two distinct modalities where the switching between the modalities is controlled by manually programmed rules [5].

The systems of the first kind require enormous manual effort to develop and maintain dialog trees for specific domains and the dialog flow configurations may not be easily used as a ground for programmatic extension of them for novel cases. In turn, the systems of the second kind may be not inspected and verified by humans and the only way to have them improved is re-training or extended loads of data.

That leads to the fact that many successful dialog systems implementing conversational intelligence are either serving narrow domains where manual effort for dialog flow configurations is economically justified or have low quality and don't scale well.

Given the amount of manual effort for configuration, maintenance, and tuning the training and re-training processes, systems of both kinds can not be called truly artificially generally intelligent, because they are not capable of extensibility and autonomous incremental learning.

Another problem acknowledged since the successful completion of the classic Turing test by "Eugene Goostman" chatbot, implementing dialog flow approach mentioned above, is lack of a measurable criteria of intelligence to be achieved by a conversational intelligence system pretending to be artificially generally intelligent, capable to evolve, learn and adapt incrementally through the lifetime.

From a practical standpoint, all of the mentioned problems have been faced in our Aigents® project [6] which needs a conversational interface for a user to be evolved for any subject domain incrementally in course of the interaction with the users themselves, so the artificial conversational intelligence agent can master its dialog skills and scope of knowledge incrementally. Not having all of the indicated problems solved yet, we approach them in a simplified form using synthetic controlled language [7,8] and having the process of language and skills acquisition under the framework relying on incremental version of the Turing test called "baby Turing Test" [9].

## 2      Controlled language as an interlingua

The idea of using controlled or synthetic language (say "artificially natural language") for human-machine interaction origins from a suggestion made by Ben Goertzel to employ synthetic Lojban language for the purpose [10] and our later development of synthetic Aigents Language [7,8] for the same goal. The difference between the two is that Lojban appears more complex resembling the complexity of a real human language to a greater extent, not being similar to any of them. On the other hand, the Aigents Language is more simple and more well-structured, inheriting most of its grammar from Turtle language used for semantic web programming [11]. Also, the syntax of the latter can be loaded with a lexicon of any human language such as English.

The grammatical forms of the controlled language and its vocabulary are strictly limited, avoiding homonymy and ambiguity. That makes named entity extraction and relationship extraction much more simple than in the case of conventional natural languages. The vocabulary of such language on itself can be extended but the rules of its extension are pre-defined as part of the language grammar and its core lexicon.

What is called controlled language here is known in the industry as domain-specific natural language dialects such as Air Traffic Control Language [12], where strictly limited set of grammatical constructs and lexicon entries make communications between pilots and air traffic controllers compact, unambiguous and efficient. Another somewhat opposing examples are local dialects of known European languages with stripped grammars and lexicons called "pidgins" [13], such as ones used in former colonies or spots of dense international tourism traffic. Such pidgins are used for cross-national spoken interchanges between carriers of different mother tongues relying on some major "carrier" language. Typically, the stripped and over-simplified versions of English are used for the purpose.

In our work [6,7,8] we use this kind of controlled language or pidgin to experiment with incremental acquisition of conversational intelligence skills and use this language in production for interaction between users and the system in chat mode.

## 3  Baby Turing test for conversational intelligence

While the success of "Eugene Goostman" chatbot passing the classic Turing test is claimed to be misleading by researchers [14], the real alternative for the test suitable for artificial general intelligence purposes has been suggested 35 years ago by Barbara Partee [9] and worth full quoting: "What the linguist demands is a "child" machine that starts with no particular language and succeeds in learning whatever language the adult humans (or machines) around it are speaking, with no more explicit instruction than a human child gets. We could call this the "baby Turing test" for linguistic ability. Distant as such a goal may seem, its recognition as a goal, at least metaphorically, could fairly be said to have marked the beginning of the tremendously explosive progress we have seen in generative grammar in the last several decades. Linguists are primarily interested in human abilities, not machine capabilities, but insofar as they demand of their theories this kind of explicitness, a successful theory should be able to help AI researchers get their machines to pass the Turing test (and the harder baby Turing test)."

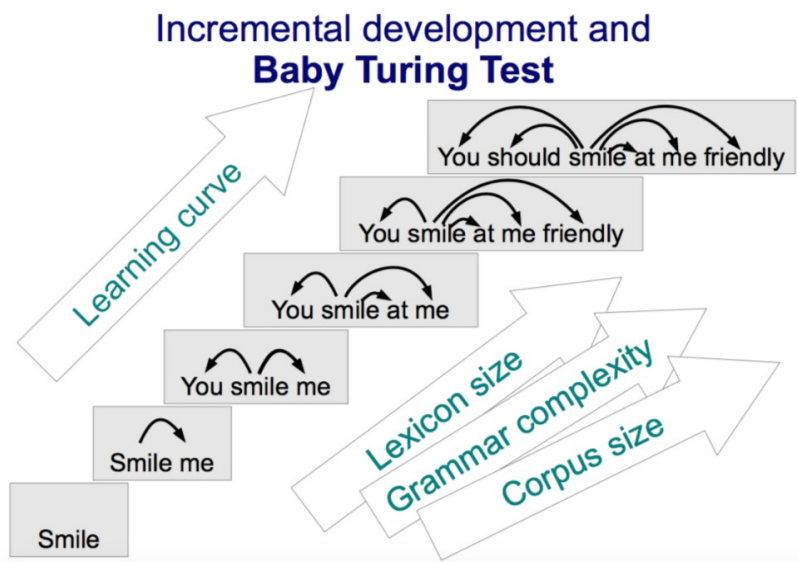

**Fig. 1.** Baby Turing Test idea is seen as an incremental acquisition of conversational skills along with the learning curve from simple nouns and pronouns to verbs in non-transitive form, then verbs in a transitive form then adding adjectives, etc. - increasing size of lexicon and complexity of grammar along with a gradual increasing of training corpus size during the life experience.

The baby Turing test reflects the incremental nature of development for an intelligent being. The nature of the test is that you take the newborn "black box", capable of nothing, then feed it with data, knowledge, and your parental feedback. You increase the complexity of your inputs gradually as shown in Fig.1. In the end, the "black box" should be able to render intelligent behavior such as required by standard Turing Test. It is well matching the entire idea of artificial general intelligence as the capability to learn new skills adapting to changes in the environment [15].

Given the modern approach to software development and testing methodology [16] involving suites of automated regression and functional testing, we can imagine an automated testing framework with an incremental testing setup where a change in system responses based on increasingly more complicated requests is verified against the expectations based on the expected learning curve trajectory.

## 4 Aigents Language as an extensible graph manipulation language with human-friendly syntax

Our work on Aigents® project [6,7,8] involves all of the aspects described above. Since our ultimate goal is to create an artificial agent capable of online information processing on behalf of its user, we need it to be able to acquire communication skills for any subject domain that the user may be interested of. We are targeting to create agent capable of general conversation intelligence with the level of intelligence increasingly growing in the course of interaction.

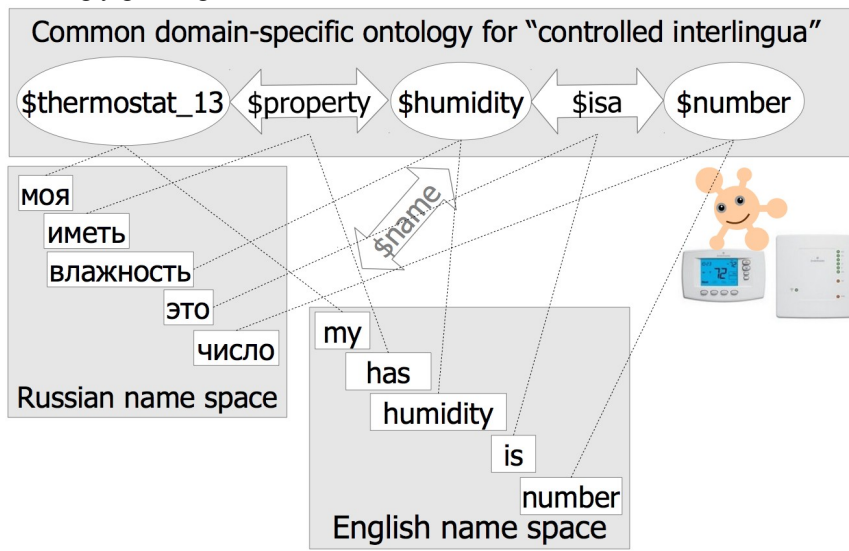

**Fig. 2.** Aigents Language as a graph manipulation semantics represented in natural language syntax where the same semantic underlying subgraph can be mapped to different namespaces with corresponding human language specific vocabularies.

To make the problem less challenging from the very beginning, we have decided to replace the English language with surrogate Aigents Language (AL) in the conversational interface to our system [7,8]. Indeed, the recent discoveries [17] made in the course of OpenCog unsupervised learning project show that the neuro-symbolic framework for unsupervised language acquisition provides better results on controlled corpora encompassing "small world" semantics and vocabulary, compared to larger literary corpora.

The specification for the language is given on or earlier works [7,8] but there are few important language features to have outlined here. It can be seen as a graph manipulation language like Turtle [11], as shown in Fig. 2. However, AL uses natural language references instead of identifiers so it can be consumed and authored by humans without the need to refer to the glossary of identifiers. Moreover, the references on themselves are enclosing declarative statements identifying the subjects of the references. In many cases, when a name of an object can be used as a unique identifier for an object being referenced, the implicit reference by name may be employed. For instance, the statement "(name Alan, surname Turing) (name is) (name scientist)" can be rewritten as "(name Alan, surname Turing) is scientist" in case if words "is" and "scientist" are non-ambiguous so they don't have homonyms. Moreover, if there is only one person with the name "Alan" in the current system state, it can be rewritten as "Alan is scientist".

Since terms in AL are not words but a *things* (i.e. concepts) behind the scene, the statement carrying the same semantics can be transparently transformed from a namespace with names in one language to the other namespace, like the statement "Alan is scientist" in English namespace could appear as "Алан это учёный" in Russian namespace.

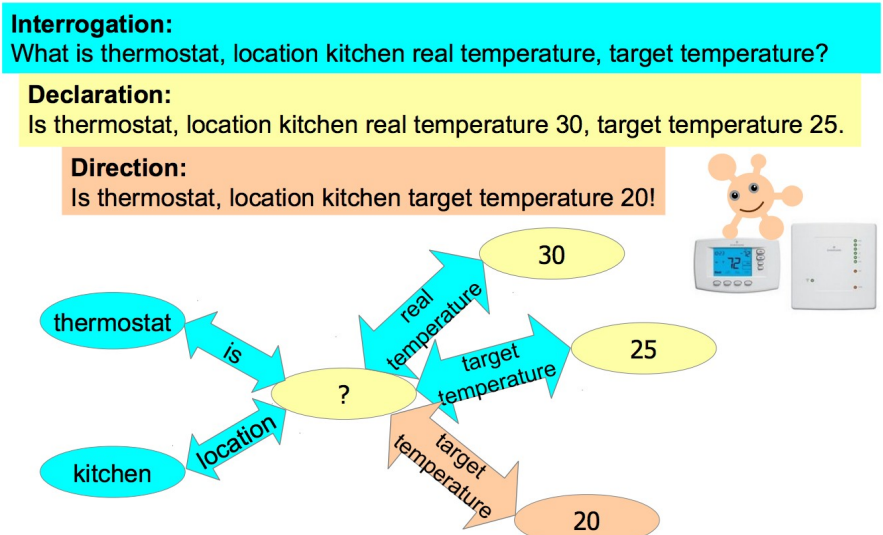

**Fig. 3.** Aigents Language encompassing (top to down): interrogative statements; replies to them as declarative statements; orders as imperative statements – all under the same semantic graph pattern using unified syntax.

Example on Fig. 3 presents communication with a "Smart House" controller. Interrogation has labeled links "is" and "location" are used to locate the object being queried, setting property values for these links as nodes named "thermostat" and "kitchen", respectively, while "temperature" links denote properties being queried for the target object. Declarative response fills those properties with nodes corresponding to an actual temperature value. Imperative order re-connects the target object's "target temperature" property with a new value to be set by the order.

Another important feature of the AL is its built-in capability to handle symmetric and asynchronous interactions between the two communication ends, unlike the SQL or GraphQL, where the query syntax is used for requests from client to server while plain text or RDF data are submitted back from server to client. The AL assumes the response may contain query itself as a reference with the declarative statement or be another query itself. This makes it similar to Atomese language employed in the OpenCog system [18].

Finally, the AL comes with minimalistic foundation ontology, which consists of basic *thing* (concept), which can be acting as a class or as an instance under different circumstances, and two properties called *is, has*, and *name*. The two properties can use used to engineer any inheritance, instantiation, and possession relationships between the *things*, so any domain-specific ontology can be engineered on the fly. In turn, the *name* property can be used to expose the ontologies to the conversational interface, having most of the terms implicitly referenced by words in the statements involved in conversation flow.

Over the last 5 years, the AL serves as a ground for Aigents® system development involving different aspects of online information search, classification, monitoring, and extraction [1] as well as a test-bed for various experiments related to conversational intelligence.

## 5    Applying Baby Turing test approach for development of conversational interface

The traditional regression and functional testing methodology in the software industry relies on the test cases where each test case verifies system operability with respect to one specific case of functionality [16]. Each of the cases is typically executed against a cleanly initialized system (in case of integration tests) or a system unit (in case of unit tests). The tests may be automated or not but the idea assumes that the system is capable to execute verifiable behavior out of the box, being initialized with a test case setup data and executing the scenario of the case, providing expected outputs along the way – having the outputs verified by the testing framework.

Practically, the same test cases may be used for two purposes: A) regression testing, ensuring that previously existing functionality is not broken with newly introduced changes; B) functional testing in test-driven development when a test case is created before the new functionality is created so passing all test cases become criteria for satisfying functional requirements for the system. This approach can be nicely leveraged to machine learning and artificial intelligence domain assuming that functional testing corresponds to verification of the system ability to reach the baseline accuracy threshold on test data sets for new training data while regression tests correspond ability to reach the same baseline thresholds on the test data including legacy data sets, ensuring that no "catastrophic forgetting" [19] is taking place.

Once we apply the same approach to incremental learning for conversational intelligence, we discover the need to have some adjustments to the approach. That is, there is no clear difference between the test case initialization and test case scenario execution anymore. On the incremental learning curve, each previous step becomes initialization for the subsequent step, so each of the test cases would consist of steps with gradually increasing complexity with performance verified on every step, relying on the previous steps. As part of the verification, the previous steps should not just supply the data necessary to learn the subsequent steps. The previous steps should verify the lack of the skill being learned along with the test case, so the system can be verified for the capability to learn on the fly instead of being engineered to perform each of the skills in advance, like it is shown on Fig. 4.

The approach outlined above has been successfully implemented as part of Aigents development process. Each of the new skills to be acquired by the system needs two pieces: architectural changes in the cognitive architecture of the system [20] itself plus input from the outer world to unlock and train the skill, having the input coming as conversational data. In the course of the development cycle for a new system skill or product feature, the test case is written first in a form of dialog script of a user communicating with a system – starting with confirmation of the lack of the skill and then feeding the system with new data and verifying the changes in its behavior along with the dialog flow till the presence of skill is confirmed. Once the script is written, the development begins so the system internal architecture is being changed and fixed until the dialog runs as expected. Along the way, all dialogs that were successfully run in the past are being executed and verified on an ongoing basis ensuring that the new changes are not breaking the legacy functionality, till the point when all new and previously existing dialog scenarios are running as expected. That is, the automated test-driven development is naturally joined with regression testing methodology [16] applied to the development of conversational intelligence.

Since the conversational interaction with an Aigents® software agent may be performed on multiple channels including operating system console, messengers, SMS/text gateways, raw TCP/IP, or HTTP GET/POST, the test could be based on any communication media. For historical reasons, we use HTTP GET/POST as a carrier protocol for the testing framework, having about 6K lines of dialog flow lines written so far, each line representing either user request or expected system reply.

The Aigents® Social and Media Intelligence Platform is implemented in Java and available open-source on GitHub: https://github.com/aigents/aigents-java under MIT license. It provides integration of data from any static Web sites (excluding those with dynamic HTML generation) with content from social networks, messengers and blockchains such as Facebook, Facebook Messenger, Slack, Telegram, Reddit, VKontakte, Steemit, Ethereum and Golos.id, providing content aggregation, knowledge extraction and social performance analytics across all these media sources.

```
    SAY:What surname Turing firstname?
    GET:There not.
    SAY:What person has?
    GET:Person not.
    SAY:There name person.
    GET:Ok.
    SAY:Person has firstname, lastname.
    GET:Ok.
    SAY:What person has?
    GET:Person has firstname, lastname.
    SAY:There is person, firstname Alan, lastname Turing
    GET:Ok.
    SAY:What is person?
    GET:There firstname alan, is person, lastname turing.
    SAY:What lastname Turing firstname?
    GET:There firstname alan.
    SAY:What lastname Turing birth date?
    GET:There not.
    SAY:Person has birth date.
    GET:Ok.
    SAY:Firstname Alan, lastname Turing birth date 23/06/1912.
    GET:Ok.
    SAY:What person has?
    GET:Person has birth date, firstname, lastname.
    SAY:What lastname Turing birth date?
    GET:There birth date 23/06/1912.
    SAY:What is person?
    GET:There birth date 23/06/1912, firstname alan, is person,
lastname turing.
```

**Fig. 4.** Basic example of conversational test case flow with incremental knowledge acquisition along with expression with newly acquired knowledge.

The conversational intelligence testing framework, including the test execution code running over HTTP GET/POST and the tests themselves, are written in PHP and available open source in the same repository: https://github.com/aigents/aigents-java/tree/master/php/agent .

The sample output of the complete run of the testing suite can be found there as well: https://github.com/aigents/aigents-java/blob/master/test.out .

The interface to the system kernel is 100% dialog-based, so even graphical user interface layers for Web, Android, and desktop applications are running on top of the Aigents Language protocol, which requires having respective functional cases to be covered by the conversational flow cases as well.

## 6       Conclusion

The artificially designed controlled language with a syntax derived from Turtle language and resembling oversimplified "pidgin" English may be used as a primary interface for a cognitive architecture (Aigents® Social and Media Intelligence Platform in our case) to perform and control content aggregation, knowledge extraction and social performance analytics across a wide range of media sources.

The development cycle can be taken under control with the use of a testing framework relying on "baby Turing test" methodology performing both functional and regression automated tests for conversational intelligence.

Our further work will be dedicated to: 1) adding more skills to the Aigents® system for more accurate performance; 2) extending controlled "Aigents Language" conversational interface to use conventional English language.